\documentclass[journal]{IEEEtran}

\usepackage{times}
\usepackage{epsfig}
\usepackage{graphicx}
\usepackage{amsmath}
\usepackage{amssymb}
\usepackage{cite}
\usepackage{color}
\usepackage{algorithm}
\usepackage{algorithmic}

\makeatletter
\newcommand{\thickhline}{
    \noalign {\ifnum 0=`}\fi \hrule height 1pt
    \futurelet \reserved@a \@xhline
}
\newcommand{\tabincell}[2]{\begin{tabular}{@{}#1@{}}#2\end{tabular}}
\usepackage{multirow}
\usepackage[pagebackref=true,breaklinks=true,letterpaper=true,colorlinks,bookmarks=false]{hyperref}

\begin{document}

\title{Global and Local Sensitivity Guided Key Salient Object Re-augmentation for Video Saliency Detection}

\author{\
    Ziqi Zhou\textsuperscript{1},
    Zheng Wang\textsuperscript{1}$^{,\ast}$,
    Huchuan Lu \textsuperscript{2},
    Song Wang \textsuperscript{3},
    Meijun Sun\textsuperscript{1}\\
    \textsuperscript{1}{College of Intelligence and Computing, Tianjin University, Tianjin, China}\\
    \textsuperscript{2}{Dalian University of Technology, Dalian, China}\\
    \textsuperscript{3}{Department of Computer Science and Engineering, University of South Carolina, Columbia, SC, USA}\\
    \emph{\{ziqizhou, wzheng, sunmeijun\}@tju.edu.cn, lhchuan@dlut.edu.cn}\\
      \emph{songwang@cec.sc.edu}
}

\maketitle

\begin{abstract}
The existing still-static deep learning based saliency researches do not consider the weighting and highlighting of extracted features from different layers, all features contribute equally to the final saliency decision-making. Such methods always evenly detect all "potentially significant regions" and unable to highlight the key salient object, resulting in detection failure of dynamic scenes. In this paper, based on the fact that salient areas in video are relatively small and concentrated, we propose a \textbf{key salient object re-augmentation method (KSORA) using top-down semantic knowledge and bottom-up feature guidance} to improve detection accuracy in video scenes. KSORA includes two sub-modules (WFE and KOS): WFE processes local salient feature selection using bottom-up strategy, while KOS ranks each object in global fashion by top-down statistical knowledge, and chooses the most critical object area for local enhancement. The proposed KSORA can not only strengthen the saliency value of the local key salient object but also ensure global saliency consistency. Results on three benchmark datasets suggest that our model has capability of improving the detection accuracy on complex scenes. The significant performance of KSORA, with a speed of 17FPS on modern GPUs, has been verified by comparisons with other ten  state-of-the-art algorithms.
\end{abstract}

\section{Introduction} \label{Sec.01}
Visual saliency models aim to detect globally important and eye-catching regions in a scene by exploiting human visual attention system characteristics. Visual saliency detection, as an important research topic in the field of computer vision, has certain inspiration and promotion effects on the development of many visual tasks, such as person re-identification \cite{7437489, Song_2018_CVPR}, visual tracking \cite{Zhu_2018_CVPR,Wang_2018_CVPR}, video object segmentation \cite{Oh_2017_CVPR}, video Compression \cite{5223506} and video captioning \cite{8100131}. According to application scenarios, the study of saliency detection can be further divided into two branches: 1) static saliency and 2) dynamic saliency. The former detects significant objects from single image by highlighting spatial high-level semantic objects, while the dynamic saliency detection, which is also known as video attention prediction, attempts to extract spatial features, temporal continuity and motion information from consecutive video frames to identify salient objects.
\begin{figure}[t]
\begin{center}
   \includegraphics[width=1\linewidth]{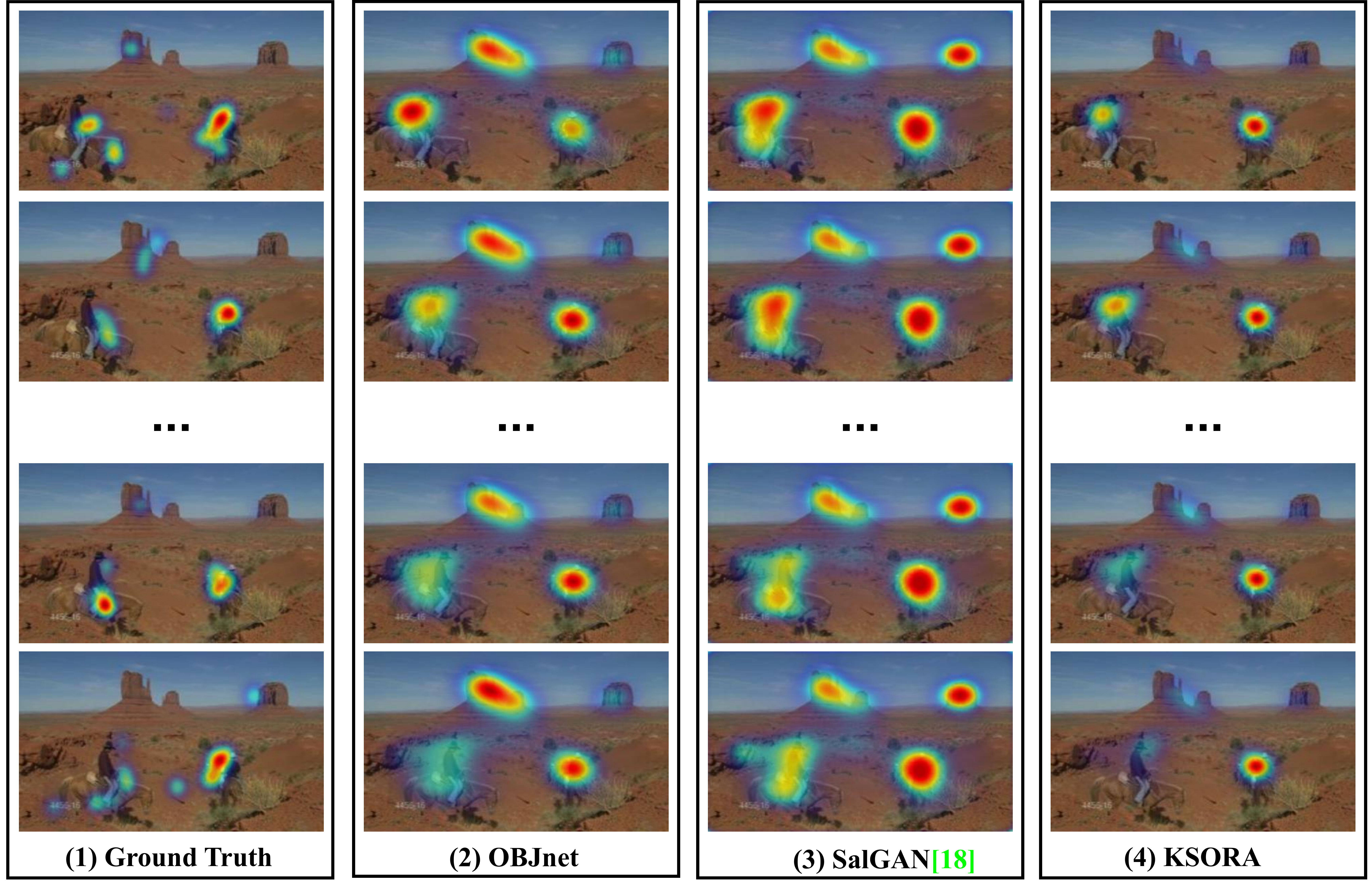}
\end{center}
   \caption{Visualization of the still-static model \emph{SalGAN}\cite{Pan_2017_SalGAN}, the baseline static model \emph{OBJnet}, and the proposed model \emph{KSORA}. It is obvious that when dealing with some complex dynamic scenes, still-static models failed to accurately locate the key salient regions, while our model can achieve accurate saliency detection.}
\label{fig:01}
\end{figure}

Early researches on saliency detection relied on hand-crafted features, such as luminance, intensity and orientation. Models such as \cite{730558} mainly use biological features to extract salient objects, while some other models \cite{harel2007graph,5963689} utilize pure mathematical calculations, such as frequency domain residuals and graph-based methods. With the development of deep learning in recent years, many researches based on auto-encoders \cite{7051244}, full convolutional neural networks (FCNNs) \cite{7488288,8237294,8382302}, and long-short term memory networks (LSTM) \cite{8400593,Gorji_2018_CVPR,Li_2018_CVPR,DBLP:journals/corr/abs-1709-06316,Ramanishka_2017_CVPR} have emerged. By extracting deep features, CNNs can achieve much better results on many public datasets than traditional methods based on hand-crafted features. However, in terms of dynamic saliency detection, state-of-the-art still-static models are unable to achieve good detection results (as shown in Figure \ref{fig:01}). This is because unlike static detectors,  which only focus on areas that are strongly contrasted with the surroundings and areas that have high-level semantic information, dynamic saliency detectors need to process continuous video data, in which moving objects, long-term appeared objects or newly emerged objects are more attractive to human attention. In addition, since the video is always played continuously, human eye fixations only stay for a short time on each frame, so the salient region on each frame is always relatively small and concentrated.
\begin{figure*}[hpt]
\begin{center}
   \includegraphics[width=\linewidth]{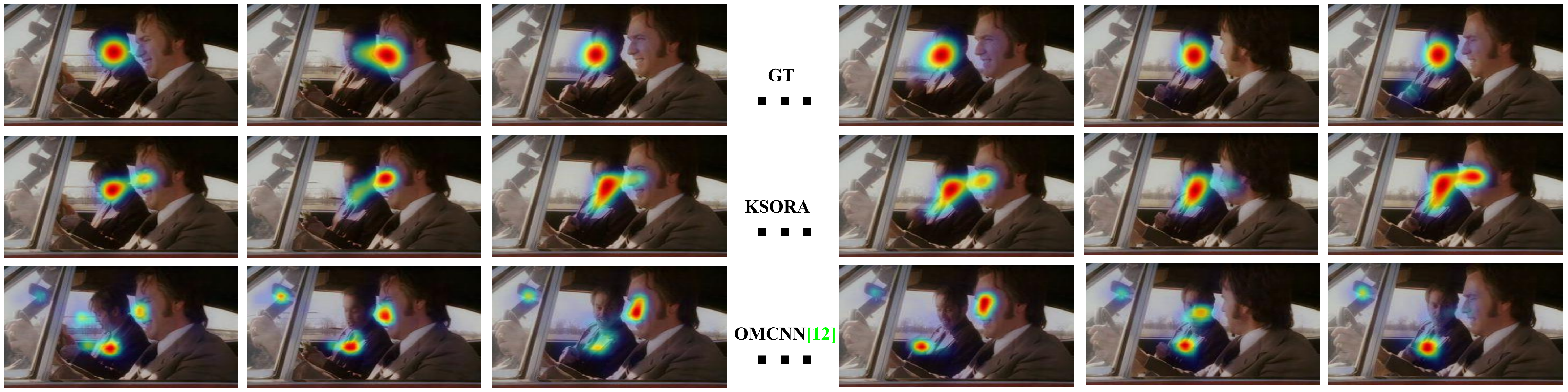}
\end{center}
   \caption{In some cases, dynamic model \emph{OMCNN} \cite{DBLP:journals/corr/abs-1709-06316} (third row) lost the key salient object, but our \emph{KSORA}(second row) always maintains the precise positioning of the key salient object, which prove the validity of our proposed model. The first row represent human fixation.}
\label{fig:02}
\end{figure*}

According to the statistical knowledge of existing video saliency datasets and researches on the human visual attention mechanism, we found that, a) moving objects are more attractive than static ones; b) objects that appear for a long time are more likely to get the most attention in the next frame at a higher probability; c) newly emerged objects are more noticeable than old ones; d) when there is no dominant object or the dominant object disappears, human attention will be refocused on the center of the screen. Based on these facts, the core problem to be solved in the study of dynamic saliency detection lies in the selection of key salient object, instead of extracting all potential salient areas equally as static saliency detection methods. In order to achieve the selection of key salient object, we need to : \textbf{i)} extract sufficient intra-frame spatial features and inter-frame motion features; \textbf{ii)} mine the associations within the features and the relationships between objects to re-augment the key salient object features for subsequent saliency inference.

In this paper, we develop two sub-modules for the selection and re-augmentation of key salient object: the weighted feature extraction module \emph{WFE} and the key object selection module \emph{KOS}. In particular, \textbf{1)} \emph{WFE} accepts the original intra-frame object features and  inter-frame motion features as input and outputs the corresponding weight response map with the same size as the input feature by adaptively learning and adjusting the internal weights. Through the adjustment of the weight response map, the features are screened at the feature level; \textbf{2)} \emph{KOS} ranks objects according to the statistical information of the current video at advanced semantical level. Its purpose is to select the most critical object area in current frame and strengthen it locally, so as to enhance the difference between the significant object and the surroundings. Some performance visualizations are shown in Figure \ref{fig:02}, our model achieves significant superiority in terms of performance when compared with state-of-the-art model \emph{OMCNN}\cite{DBLP:journals/corr/abs-1709-06316}. The fixation prediction result of \emph{KSORA} is more accurate on locating the key object and more in line with the realistic statistic probability distribution.

The contribution of this work is threefold:

\textbf{(1)} we propose a key salient object re-augmentation method \emph{KSORA} based on top-down semantic knowledge and bottom-up feature guidance to improve detection accuracy in video scenes;

\textbf{(2)} we design an weighted feature extraction module \emph{WFE} for feature level saliency decision and then perform feature pyramid fusion for sufficient information extraction.

\textbf{(3)} we introduce a key object selection module \emph{KOS} to enhance key salient object from high-level semantics, and to re-adjust the saliency probability distribution to ensure that the saliency probability of the key salient object remains at the leading level.

The remainder of this paper is organized as follows: in Section \ref{Sec.02}, we briefly survey several related works. In Section \ref{Sec.03}, we introduce the entire structure and each sub-module of the proposed model \emph{KSORA}. Experimental results are outlined in Section \ref{Sec.04}, and Section \ref{Sec.05} concludes the paper.
\begin{figure*}[hpt]
\begin{center}
    \includegraphics[width=1\linewidth]{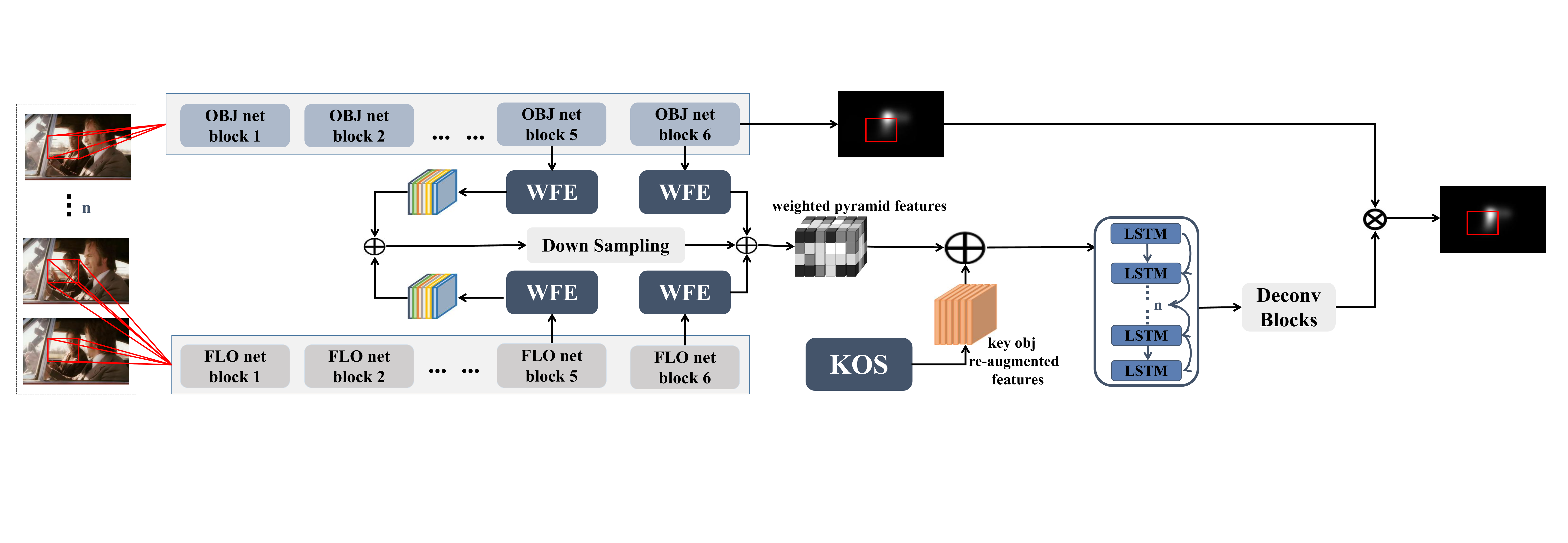}
\end{center}
   \caption{ The overall structure of the proposed model \emph{KSORA}. As shown in the figure, the baseline network is \emph{OBJnet}. \emph{FLOnet} is employed for motion feature extraction, \emph{WFE} is designed for bottom-up weighted pyramid feature calculation and \emph{KOS} is for top-down key salient object selection. Architecture of \emph{WFE} and \emph{KOS} will be introduced in detail in Figure \ref{fig:04}.}
\label{fig:03}
\end{figure*}
\section{Related Works} \label{Sec.02}
In this section, we briefly introduce the works related to the proposed.
\subsection{Static Saliency Detection}
The static, still-image saliency detection approaches have been deeply studied for years, which can be divided into bottom-up low-level feature extraction based methods \cite{7051244,8382302,7488288,Wang_Lu_2018_CVPR,Zhang_2018_CVPR,8400593,8237294,Pan_2017_SalGAN,harel2007graph,730558,Wang_2017_CVPR} and top-down high-level semantic knowledge guided methods \cite{Ramanishka_2017_CVPR,7442536,6751329}. Representatively, \cite{harel2007graph} proposed a graph-based algorithm, which still used the feature extraction method of \cite{730558}, and used Markov chain to calculate the interaction of "center-surround" pixel patches. \cite{8382302,7488288,Wang_Lu_2018_CVPR,Zhang_2018_CVPR,8237294} proposed FCNs-based end-to-end detection methods, in which \cite{8382302,7488288} paralleled the saliency detection task to the image segmentation task, the training process of the model is performed in a multi-task manner; \cite{Zhang_2018_CVPR,8237294} improved the effect of the FCNs-based model by proposing multi-scale feature extraction strategies, \cite{Wang_Lu_2018_CVPR} localized salient objects more accurately by using weighted response map. \cite{Wang_2017_CVPR} proposed a weak supervision training method, which got rid of the dependence on pixel-level labels. The model used image-level labels for pre-training, and then optimized itself by iterative enhancement. \cite{Pan_2017_SalGAN} proposed a GAN-based image saliency detection method. The generator model was continuously optimized under the supervision of the discriminator, resulting in a smoother edge of the saliency map. \cite{8400593} employed convLSTM as an iterative optimization part to capture spatial saliency region. By studying the captioning model, \cite{Ramanishka_2017_CVPR} learned the visual saliency guided by captions and explicitly exposed the region-to-word mapping in modern encoder-decoder networks.
\subsection{Dynamic Saliency Detection}
\begin{figure*}[hpt]
\begin{center}
    \includegraphics[width=1\linewidth]{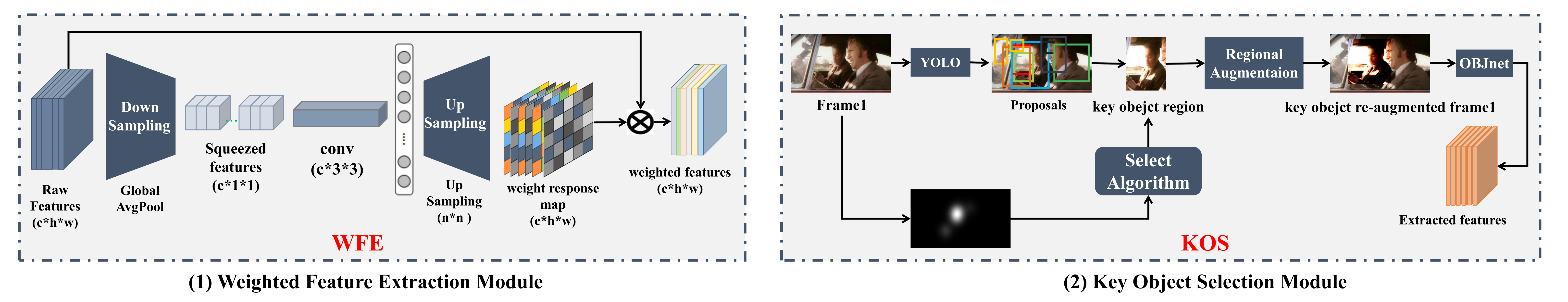}
\end{center}
   \caption{ (1) Structure of \emph{WFE}, which is used for bottom-up feature selection and feature weighting; (2) The execution details of \emph{KOS}, whose role is to rank the objects through top-down statistical knowledge and to re-extract features after key salient object region enhancement. The key salient object selection algorithm will be introduced in section \ref{subsec.02}}
\label{fig:04}
\end{figure*}
In recent years, the detection of salient regions in dynamic scenes \cite{8365810,8237450,Gorji_2018_CVPR,Li_2018_CVPR,7752954,8047320,6618996,DBLP:journals/corr/abs-1709-06316,7164324,Wang_2018_CVPR_Revisiting} can be roughly divided into two branches: 1) one performing end-to-end processing method to learn the dynamic correlation in the video sequence by adjusting the input data of the model, such as \emph{SG-FCN} model \cite{8365810}. which obtains the saliency map of the current frame directly by taking the motion boundary map of the current frame, the previous saliency map and the current frame as input. \cite{8047320} extracted the saliency map of the first frame and the subsequent $n$ frames respectively by using two different models. \cite{Wang_2018_CVPR_Revisiting} employed convLSTM to implement time continuity modeling, the feature extraction process was similar to \cite{8047320}; 2) the second branch extracts motion and object features separately and then fuse the features to get the final result. For example, \cite{DBLP:journals/corr/abs-1709-06316} used YOLO net\cite{DBLP:journals/corr/RedmonDGF15} and Flow net \cite{DBLP:journals/corr/FischerDIHHGSCB15} to extract the object features and motion features of the frame respectively, then combined the features and fed into a two-layer convLSTM module to complete the dynamic saliency prediction. \cite{Li_2018_CVPR} also designed a video saliency detection model based on similar ideas. \cite{Gorji_2018_CVPR} proposed a video saliency detection model based on multi-stream convLSTM. The model was implemented by a multi-path structure, including a saliency pathway for video saliency detection and three attentional push-pathways providing attention-driven information which mainly came from actor gaze, attention rebound, and sudden scene changes. Besides, \cite{6618996} believed that the saliency area in the video is relatively small, and if the saliency was calculated on each pixel, it will cause a lot of redundancy. Therefore, the authors selected a set of candidate gaze positions and calculated the saliency only at these locations. \cite{8237450} developed a depth-aware video saliency approach to predict human focus of attention when viewing videos that contain a depth map (RGBD) on a 2D screen. \cite{7752954} achieved better detection by computing background priors.

As mentioned above, most existing spatiotemporal saliency models incorporated various motion cues into the existing static models in the literature, but as we analyzed in Section \ref{Sec.01}, in addition to motion characteristics, the appearing time of the object ( Long-term exist or new occur) also has an impact on fixation. Therefore, our starting point is how to screen and rank the salient objects. For this purpose, we propose \emph{KSORA} with two sub-modules \emph{WFE} and \emph{KOS} to improve the detection accuracy by re-enhancing the key salient object features based on the analysis of existing video saliency datasets. \emph{WFE} implements weighted feature screening using bottom-up strategy while \emph{KOS} works as an error correction function to enhance the global key salient object region using top-down semantical statistical knowledge. It is worth noting that both the two sub-modules are actually sub-branches of the entire model, so the model can be considered as an end-to-end network.

\section{Our Approach} \label{Sec.03}
In this section, we will introduce our model (\emph{KSORA}) presented in this research in more detail.
\subsection{Local Sensitivity Guided Weighted Feature Extraction}\label{subsec.01}
Quality of the extracted features is decisive and critical for final saliency prediction. For high-quality feature extraction, we need to 1) obtain sufficient features, which can be achieved by multi-scale feature extraction, 2) emphasize informative features and suppress less useful ones, which can be called as salient feature selection. As for 1), \cite{8240654} has proved that extracting multi-scale features can improve the performance of the model. It is well known that higher layer captures higher semantic information, while lower layer processes lower-level features. As multi-scale features come from different layers, the overall performance can be improved by obtaining sufficient features with different receptive field sizes. Regarding 2), although extracting sufficient features can promote the performance of the model, not all features play a positive role. Therefore, filtering and screening features to emphasise informative features while weakening useless features is also very helpful to improve the performance of the model. To the best of our knowledge, there are no proposed works filtering extraction feature when performing saliency detection, and this is the first work considering local-feature level saliency decisions.

\begin{figure}[h]
\begin{center}
   \includegraphics[width=1\linewidth]{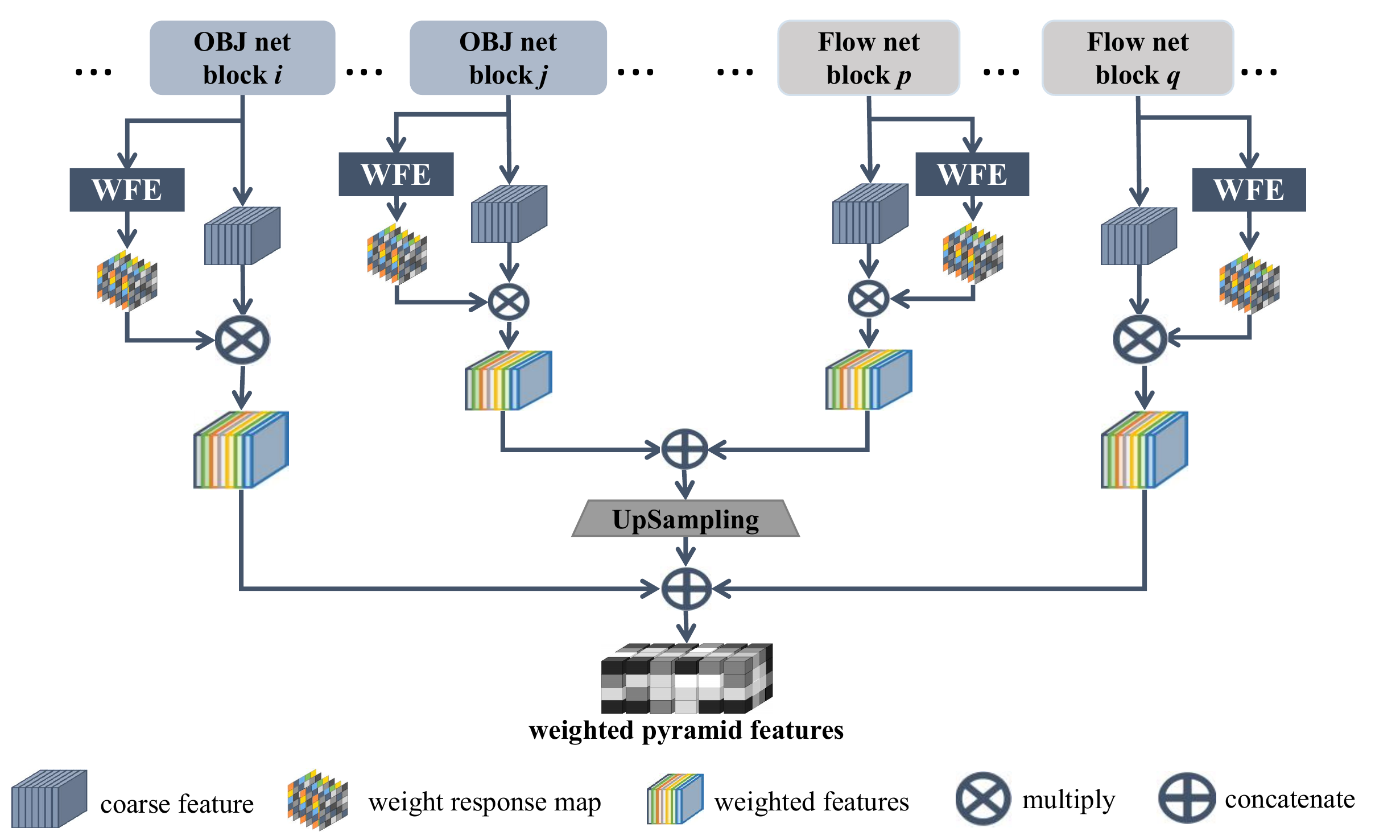}
\end{center}
   \caption{Pyramid weighted feature fusion strategy. Weighted features from different layers of \emph{OBJnet} and \emph{FLOnet} are fused for multi-scale feature extraction, which will improve the model's spatiotemporal perception ability.}
\label{fig:05}
\end{figure}

Based on the above discussion, we propose a weight feature extraction module \emph{WRM} to weight the original features to achieve salient feature selection, and we also employ pyramid feature fusion strategy to execute multi-scale feature extraction. The details of the pyramid feature fusion strategy and \emph{WRM} can be found in Figure \ref{fig:05} and Figure \ref{fig:04}-(1) respectively.

More specifically, considering the continuous motion relationship contained in video sequence, apart from designing \emph{OBJnet} for intra-frame spatial feature extraction and coarse saliency map calculation, we also employ flownet-based \cite{DBLP:journals/corr/FischerDIHHGSCB15} model \emph{FLOnet} to perform inter-frame motion information extraction. To obtain weighted features that contains sufficient information, we select the object features of the $i$-$th$ and $j$-$th$ layers from \emph{OBJnet} and the motion features of the $p$-$th$ and $q$-$th$ layers from \emph{FLOnet}. These selected features are weighted by \emph{WFE} to participate in the next feature pyramid calculation.

The role of \emph{WFE} is to select and filter the extracted original features and perform significant decision-making at the local-feature level. The entire process is implemented through a bottom-up strategy, which consists of three layers: a global average pooling layer, a convolution layer and an up-sample layer. First, we use a global average pooling layer to get the global relevant vector $X_g^{F_p}$ of the input feature $X^{F_p}(c, w, h)$.
\begin{equation}
    \begin{split}
        X_g^{F_p} &= (x_1^p, ..., x_j^p, ..., x_c^p)\\
        x_j^p &= \frac{\sum_{i=1}^{i=w\times h}x_{ji}^p}{w\times h}, j = 1,..., c
    \end{split}
\end{equation}

Next, we employ a $c$$\times$3$\times$3 convolution layer for spatial correlation computation ($c$ is the original feature channel size), the output of this layer can be:
\begin{equation}
    X_c^{F_p} = f(\Sigma X_g^{F_p} + b)
\end{equation}

Finally, we up-sample the extracted weight response map to obtain a weight matrix with the same size as the original input. We use $g$ to represent the up-sampling operation, then the weight response map $W_{X^{F_p}}$ can be expressed as:
\begin{equation}
    W_{X^{F_p}} = g(X_c^{F_p})
\end{equation}

After obtaining the weight response map $W_{X^{F_p}}$, the weighted feature $X_w^{F_p}$ can be obtained by:
\begin{equation}
   X_w^{F_p} = X^{F_p} \odot W_{X^{F_p}}
\end{equation}
where $\odot$ is the pixel-wise multiplication. It is worth noting that our method which not only considers channel-wise relationship but also takes intra- feature map spatial dependencies into account, is different from \cite{Hu_2018_CVPR}. \emph{WFE} can adaptively strengthen channel-wise salient features and emphasize key information inside the feature map while preserving the spatial structure of the features.

Once weighted the original features from several layers, we fuse these weighted features to obtain the feature pyramid for subsequent saliency inference, as shown in Figure \ref{fig:05}. The final feature pyramid $\Gamma$ can be represented as:
\begin{equation}
    \Gamma = \zeta(\mathit{X}_w^{F_p}, \mathit{X}_w^{O_i}, \phi(\zeta(\mathit{X}_w^{F_q}, \mathit{X}_w^{O_j})))
\end{equation}
where $\mathit{X}_w^{O_i}$ and $\mathit{X}_w^{O_j}$ are weighted features from \emph{OBJnet}, $\mathit{X}_w^{F_p}$ and $\mathit{X}_w^{F_q}$ are weighted features from \emph{FLOnet}, $\zeta$ represents channel-wise addition, and $\phi$ represents down-sampling.

Subsequently, we screen the original inter-frame motion features and intra-frame object features to prepare for subsequent saliency predictions.
\subsection{Global Sensitivity Guided Key Object Selection}\label{subsec.02}
From the study of the human visual attention mechanism and the analysis of existing eye fixation datasets, we find that in the dynamic scene, three types of objects have the greatest probability of attracting human attention, including 1) objects that appear in the scene for a long time; 2) objects that carry motion information and 3) objects that suddenly appear. Firstly, the first kind of objects is temporally critical since they are likely to continue to appear in the next frame. Secondly, carrying motion information or not is an important difference between a dynamic scene and a static scene. The moving object is the main area that attracts attention to the human eye. Finally, the suddenly appearing objects break the balance that existed before, and new things are more likely to attract human attention. Therefore, a key object selection module \emph{KOS} is proposed to simulate this mechanism.

As shown in Figure \ref{fig:04}-(2), the \emph{KOS} module ranks the objects on current $t$-$th$ frame $F_t$  by statistical knowledge of the semantic level and selects the key saliency object, then the selected region is locally strengthened as the region of interest. The enhanced frame $F_{aug}^t$ will be sent to \emph{OBJnet} for feature extraction. The new extracted feature $fea_{kso}$ will be used to assist the model in performing subsequent saliency detection to achieve re-enhancement of the key salient object.

The core of \emph{KOS} is how to obtain the region of interest. When making the selection of region of interest, we should rely on the statistics of the current video, because videos are context-sensitive, meaning that the information in the time series is contextually related, so the selection of the region of interest of the current frame $F_t$ should be dependent on the statistics knowledge of previous $t$-1 frames. To handle this problem, we design a selection algorithm to make corresponding judgments for different situations. The pseudo code of the selection algorithm is shown in Algorithm \ref{alg:01}, whose purpose is to judge the significant confidence of different objects and select the region of the object with the highest confidence as the region of interest. In order to accurately calculate the significant confidence of each object, we divide the objects in the scene into three categories: 1) long-term appeared objects; 2) newly-emerged objects; and 3) detection failure objects. For these three different situations, three different ways are used to calculate the significant confidence: ($GSC$ and $GSS$ are the ranked salient object classes and scores calculated by Algorithm 2, respectively.)

i) If $GSC$ obtained is an empty set, which means that the current video scene is quite complicated and there is no particularly prominent object. Under this circumstances, the observer's attention should be more inclined to land at the center of the screen, so we employ (3, 3) Gaussian prior to smooth the whole frame and use the centra-highlighted whole frame as the region of interest;

ii) If the category of candidate region $B_s$ in the current frame is not yet present in $GSC$, it indicates that this kind of objects is a new object that suddenly appears, and the attention degree of this object should be high, so we enhance the confidence value of $B_s$ (doubling);

iii) If the category of $B_s$ is already existing in $GSC$, we calculate confidence value of $B_s$ based on the ranking of the category of $B_s$ in $GSC$ and the average attention confidence value of this class in $GSS$.

After considering these three cases comprehensively, we sort the confidence obtained on all candidates and select the top-one region for local enhancement, leaving the rest of the area unchanged. Then, we feed the local enhanced frame $F_{aug}^t$ into \emph{OBJnet} for feature extraction.

Specifically, the candidate object box set $B$ of each frame is executed by \cite{DBLP:journals/corr/RedmonDGF15}. ${conf}_i$ represents the saliency confidence value of the $i$-$th$ candidate region, $\cup(x,y)$ is used to calculate the number of x(x=1) in y, $\digamma()$ represents \emph{OBJnet}, $\Lambda(roi)\leftarrow\kappa(roi, [l_{i}, h_{i}],[l_{o}, h_{o}])$ represents the local enhancement method. We adjust the contrast and brightness of the selected area to enhance the difference between $roi$ and the surroundings.

\begin{algorithm}
\caption{Selection Algorithm}
\label{alg:01}
\begin{algorithmic}
    \REQUIRE  continues frames $F_1$,...,$F_t$, coarse saliency $S_1$,...$S_t$, objects proposals $B_1$,...,$B_t$
    \ENSURE  feature $fea_{kso}$ of current frame $t$
    \STATE $GSC,GSS \leftarrow Alg\ref{alg:02}(F_{1...t-1},S_{1...t-1},B_{1...t-1})$
    \IF {$|$ \{GSC\}$|$ == 0 }
       \STATE $roi \leftarrow F_t.* GuassBlur()$
    \ELSE
        \STATE $ N \leftarrow |\{B_t\}|$
        \STATE$score \leftarrow zeros(N)$
        \FOR {each $i \in [1,N]$}
           \STATE ${conf}_i \leftarrow \cup(S_t,B_t^i.bbox)$
           \STATE ${c}_i \leftarrow B_t^i.class$
           \IF { (${c}_i \subseteq GSC$)}
                   \STATE $rank \leftarrow GSC(c_i).RankNumber$
                   \STATE $score_i \leftarrow {conf}_i * (1+\frac{{conf}_i}{GSS_{c_i}.conf}*\frac{1}{rank})$
           \ELSE
                   \STATE $score_i \leftarrow {conf}_i * 2 $
           \ENDIF
        \ENDFOR
        \STATE $[rank\_num,rank\_score] \leftarrow sort(score)$
        \STATE $roi_{Bb} \leftarrow B_t^{rank\_num(1)}$
        \STATE $roi \leftarrow F_t(roi_{Bb}(1):roi_{Bb}(3), roi_{Bb}(2):roi_{Bb}(4))$
    \ENDIF
    \STATE $\Lambda(roi)\leftarrow\kappa(roi, [l_{i}, h_{i}],[l_{o}, h_{o}])$
    \STATE $F^t_{aug} \leftarrow (F_t - roi) + \Lambda(roi)$
    \STATE $fea_{kso} \leftarrow \digamma (F^t_{aug})$
\end{algorithmic}
\end{algorithm}

\begin{table*}
\begin{center}
\begin{tabular}{c|c|c|c|c|c|c|c|c|c|c|c|c}
        \thickhline
        \multirow{2}{*}{\footnotesize{Dataset}}&\multirow{2}{*}{\footnotesize{Metric}}&\multicolumn{2}{c|}{Static Model}&\multicolumn{8}{c|}{Dynamic Model}&\multirow{2}{*}{\footnotesize{\textbf{proposed}}}
        \\ \cline{3-12}	
        &&\footnotesize{\textbf{SalGAN}}&\footnotesize{\textbf{SAM}}&\footnotesize{\textbf{GBVS}}&\footnotesize{\textbf{PQFT}}&\footnotesize{\textbf{VSOD}}&\footnotesize{\textbf{SG-FCN}}&\footnotesize{\textbf{OMCNN}}&\footnotesize{\textbf{Hou}}&\footnotesize{\textbf{Seo}}&\footnotesize{\textbf{ACL}}
        \\ \thickhline
        \multirow{4}{0.9cm}{\scriptsize{\textbf{UCF-Sports}}\\(27)}&
        \footnotesize{CC}$\uparrow$&0.535&0.529&0.441&0.197&0.475&\color{blue}{0.599}&0.548&0.334&0.331&0.446&\color{red}{0.609}
        \\ 													
        &\footnotesize{SIM}$\uparrow$&0.336&0.403&0.250&0.172&0.358&\color{blue}{0.452}&0.408&0.291&0.251&0.325&\color{red}{0.490}
        \\ 										
        &\footnotesize{EMD}$\downarrow$&1.200&0.954&1.335&1.888&1.166&\color{blue}{0.842}&0.953&1.444&1.582&0.998&\color{red}{0.747}
        \\ 		
        &\scriptsize{AUC-J}$\uparrow$&0.878&0.876&0.867&0.729&0.792&\color{blue}{0.901}&0.881&0.814&0.774&0.854&\color{red}{0.904}
        \\ \thickhline  	     																			
        \multirow{4}{0.9cm}{\scriptsize{\textbf{HOLLY WOOD2}}\\(442)}&\footnotesize{CC}$\uparrow$&0.556&0.535&0.347&0.144&0.327&\color{blue}{0.593}&0.509&0.177&0.159&0.560&\color{red}{0.606}
        \\							
        &\footnotesize{SIM}$\uparrow$&0.425&0.467&0.304&0.216&0.315&\color{blue}{0.499}&0.428&0.242&0.238&0.439&\color{red}{0.508}
        \\	        	
        &\footnotesize{EMD}$\downarrow$&1.206&0.983&1.458&1.920&1.525&\color{blue}{0.919}&1.110&1.735&1.766&1.038&\color{red}{0.861}
        \\ 				
        &\scriptsize{AUC-J}$\uparrow$&0.849&0.858&0.839&0.639&0.760&\color{blue}{0.872}&0.832&0.704&0.676&0.852&\color{red}{0.900}
        \\ \thickhline						                                     										
        \multirow{4}{0.9cm}{\scriptsize{\textbf{LEDOV}}\\(41)}&\footnotesize{CC}$\uparrow$&0.439&0.414&0.275&0.142&0.296&0.482&\color{red}{0.558}&0.202&0.169&0.328&\color{blue}{0.496}
        \\        						
        &\footnotesize{SIM}$\uparrow$&0.263&0.305&0.153&0.115&0.224&\color{blue}{0.349}&0.347&0.177&0.154&0.215&\color{red}{0.371}
        \\
        &\footnotesize{EMD}$\downarrow$&1.094&0.891&1.366&1.836&1.390&0.834&\color{blue}{0.825}&1.493&1.580&1.068&\color{red}{0.824}
        \\ 							
        &\scriptsize{AUC-J}$\uparrow$&0.882&0.865&0.845&0.728&0.800&0.884&\color{red}{0.910}&0.791&0.756&0.844&\color{blue}{0.899}
        \\ \thickhline
\end{tabular}
\end{center}
\caption{Performance comparison of ten state-of-the-arts and the proposed model \emph{KSORA}. The performance of our model on all three datasets has remained at an advanced level, which surpass most existing state-of-the-art. The top two results are shown in {\color{red}{red}} and {\color{blue}{blue}}, respectively.}
\label{Tab:01}
\end{table*}
\begin{algorithm}
\caption{Global object Ranking}
\label{alg:02}
\begin{algorithmic}
    \REQUIRE frames $F_1$,...,$F_{t-1}$, coarse saliency $S_1$,...$S_{t-1}$, proposals $B_1$,...,$B_{t-1}$
    \ENSURE  global object class ranking number $GSC$ and score $GSS$
    \STATE $P \leftarrow \{\}$
    \FOR {each $ k \in [1,t-1]$}
        \STATE $N \leftarrow |\{B_k\}|$
        \FOR {each $j \in [1,N]$}
            \STATE ${conf}_j \leftarrow \cup(S_k,B_k^j.bbox)$
            \STATE ${c}_j \leftarrow B_k^j.class$
            \IF{ (${c}_j \subseteq P$)}
                \STATE $P{{c}_j} += {conf}_j$
            \ELSE
                \STATE $add(c_j) \Rightarrow P$
                \STATE $P{{c}_j} = {conf}_j$
           \ENDIF
        \ENDFOR
    \ENDFOR
    \STATE $ GSS \leftarrow GSS/(t-1)$
    \STATE $[GSC, GSS] = sort(P)$
\end{algorithmic}
\end{algorithm}
\subsection{Saliency Inference}
After bottom-up weighted feature extraction and top-down key object selection, we employ a two-layer convLSTM module to achieve the final saliency inference. First, we combine the weighted feature pyramid $\Gamma$ with the key salient object re-augmented feature $fea_{kso}$, and then we input these features into two-layer convLSTM. Each layer uses 0.2 dropout to improve the generalization ability and to prevent over-fitting. Specifically, for each convLSTM layer, the state transition process of its inside three gates (\emph{input}, \emph{output} and \emph{forget}) can be expressed as (assuming the input feature is $\chi_t$ at time step $t$):
\begin{equation}
    \begin{split}
        i_t&=\sigma(W_{xi}*\chi_t+W_{hi}*H_{t-1}+W_{ci}\circ C_{t-1}+b_i)\\
        f_t&=\sigma(W_{xf}*\chi_t+W_{hf}*H_{t-1}+W_{cf}\circ C_{t-1}+b_f)\\
        o_t&=\sigma(W_{xo}*\chi_t+W_{ho}*H_{t-1}+W_{co}\circ C_t+b_o)\\
        C_t&=f_t \circ C_{t-1}+ i_t \circ tanh(W_{xc}*\chi_t+W_{hc}*H_{t-1}+b_c)\\
        H_t&=o_t\circ tanh(C_t)
    \end{split}
\end{equation}
Where $i_t, f_t, o_t,H_{t-1},\chi_t$ and $C_{t-1}$ are all three-dimensional tensors, the first one represents time, and the latter two dimensions represent spatial dimensions. $i_t, f_t, o_t$ are the three gates, $c$ is the memory cell, $H$ is the hidden state. $*$ represents the convolution operation, and $\circ$ represents the Hadamard multiplication.
\section{Experiments}\label{Sec.04}
\subsection{Experimental Config}
\subsubsection{Datasets}

\begin{table}
\begin{center}
\begin{tabular}{ccccc}
        \thickhline
        \small{\textbf{Dataset}}&\small{\textbf{Resolution}}&\small{\textbf{Objects}}&\small{\textbf{Videos}}&\small{\textbf{Mode}}\\
        \hline
        \hline
        \tabincell{c}{\footnotesize{Holly}\\\footnotesize{wood2}}&\tabincell{c}{\footnotesize{528*224}\\\footnotesize{-720*528}}&\tabincell{c}{\footnotesize{Human}}&\tabincell{c}{\footnotesize{Train:823}\\\footnotesize{Val:442}\\\footnotesize{Test:442}}&\tabincell{c}{\footnotesize{Task}}\\
        \hline
        \tabincell{c}{\footnotesize{UCF}\\\footnotesize{-Sports}}&\tabincell{c}{\footnotesize{480*360}\\\footnotesize{-720*576}}&\tabincell{c}{\footnotesize{Human}}&\tabincell{c}{\footnotesize{Train:38}\\\footnotesize{Val:27}\\\footnotesize{Test:27}}&\tabincell{c}{\footnotesize{Task}}\\
        \hline
        \tabincell{c}{\footnotesize{LEDOV}}&\tabincell{c}{$\geq$\footnotesize{720p}}&\tabincell{c}{\footnotesize{Human,}\\\footnotesize{Animal,}\\\footnotesize{Man-made}\\\footnotesize{Object}}&\tabincell{c}{\footnotesize{Train:436}\\\footnotesize{Val:41}\\\footnotesize{Test:41}}&\tabincell{c}{\footnotesize{Free}}\\
        \thickhline
\end{tabular}
\end{center}
\caption{Details of three video eye tracking datasets.}
\label{Tab:02}
\end{table}
We carry out model performance evaluations on three largest common video eye tracking datasets, which include hollywood2 \cite{6942210}, ucf-sports \cite{Soomro2014Action}, and ledov \cite{DBLP:journals/corr/abs-1709-06316}. Details of the datasets and the settings for training are shown in Table \ref{Tab:02}.
\subsubsection{Metrics}
To fully evaluate our proposed model, five commonly used evaluation criteria, including Correlation Coefficient (CC), Similarity (SIM), Earth Mover`s Distance (EMD), AUC-Judd (AUC-J) and Precision-Recall (PR) are used to reflect the model performance.
\subsection{Experimental Results}
\begin{figure*}[hpt]
\begin{center}
   \includegraphics[width=\linewidth]{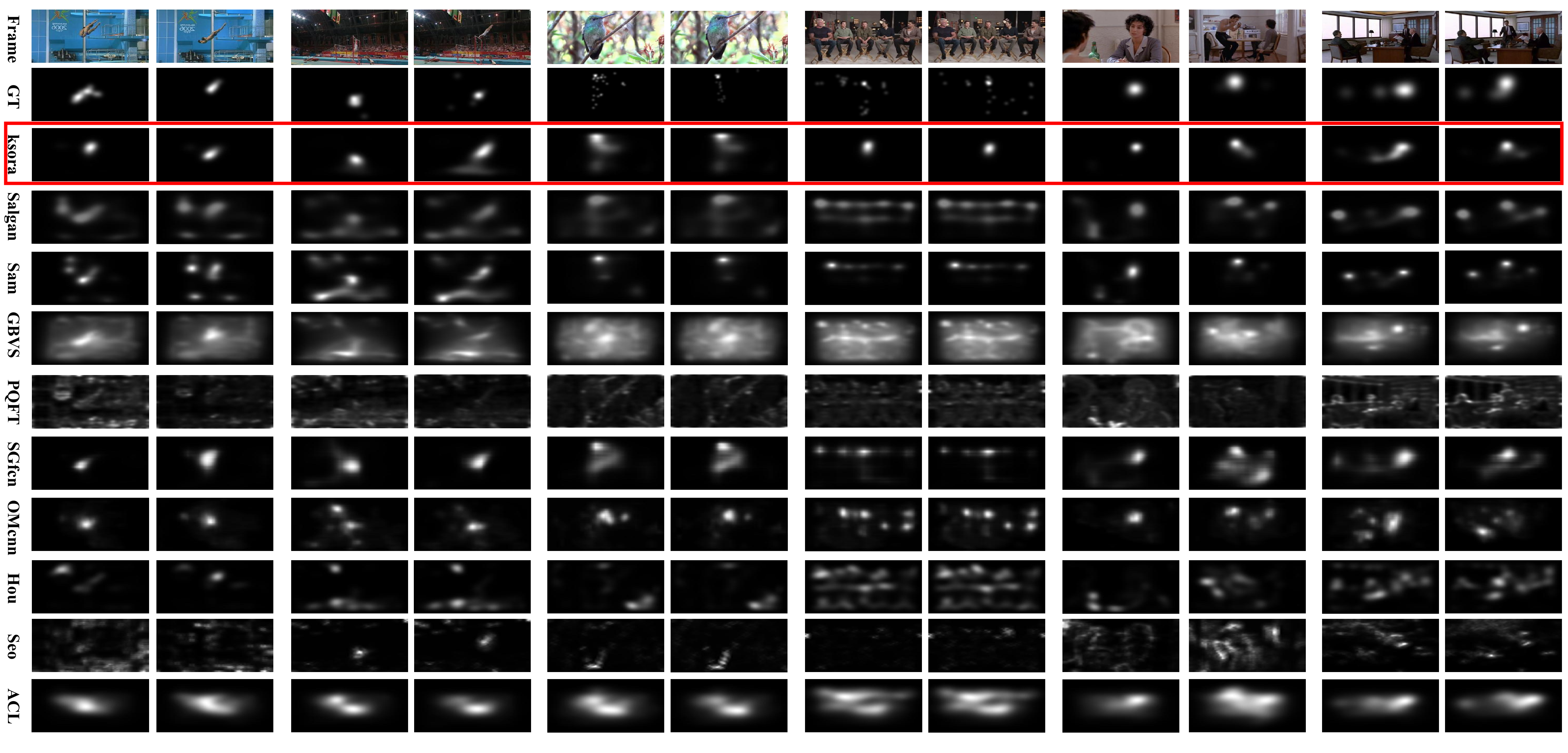}
\end{center}
   \caption{Model performance comparison with SalGAN \cite{Pan_2017_SalGAN}, SAM \cite{8400593}, GBVS \cite{harel2007graph}, PQFT \cite{5223506}, VSOD \cite{8047320}, SG-FCN \cite{8365810}, OMCNN\cite{DBLP:journals/corr/abs-1709-06316}, Hou,$et.al$ \cite{NIPS2008_3531}, Seo, $et.al$ \cite{seo2009static} and ACL \cite{Wang_2018_CVPR_Revisiting}.}
\label{fig:06}
\end{figure*}

\begin{figure}[hpt]
\begin{center}
   \includegraphics[width=0.47\linewidth]{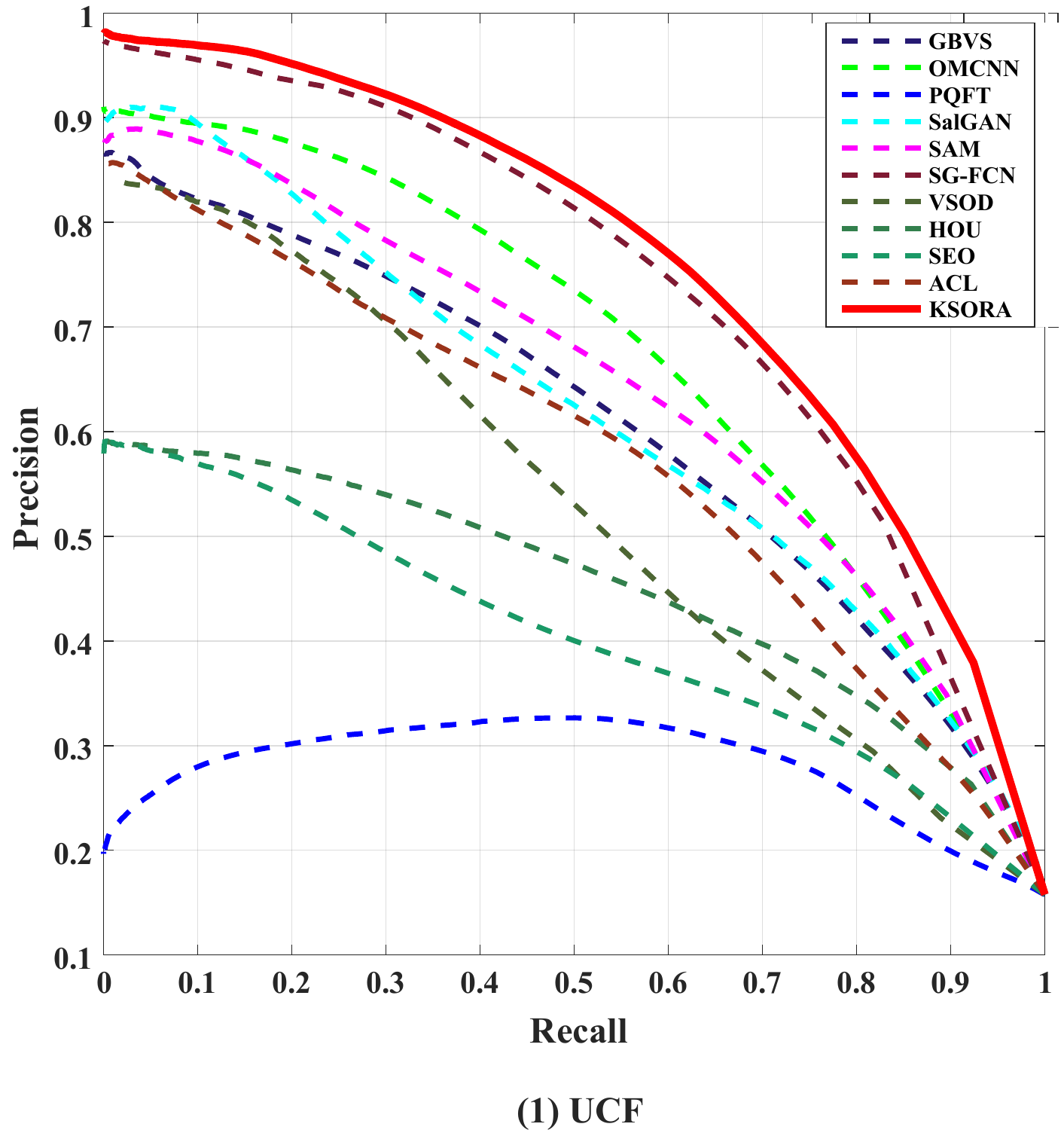}
   \includegraphics[width=0.47\linewidth]{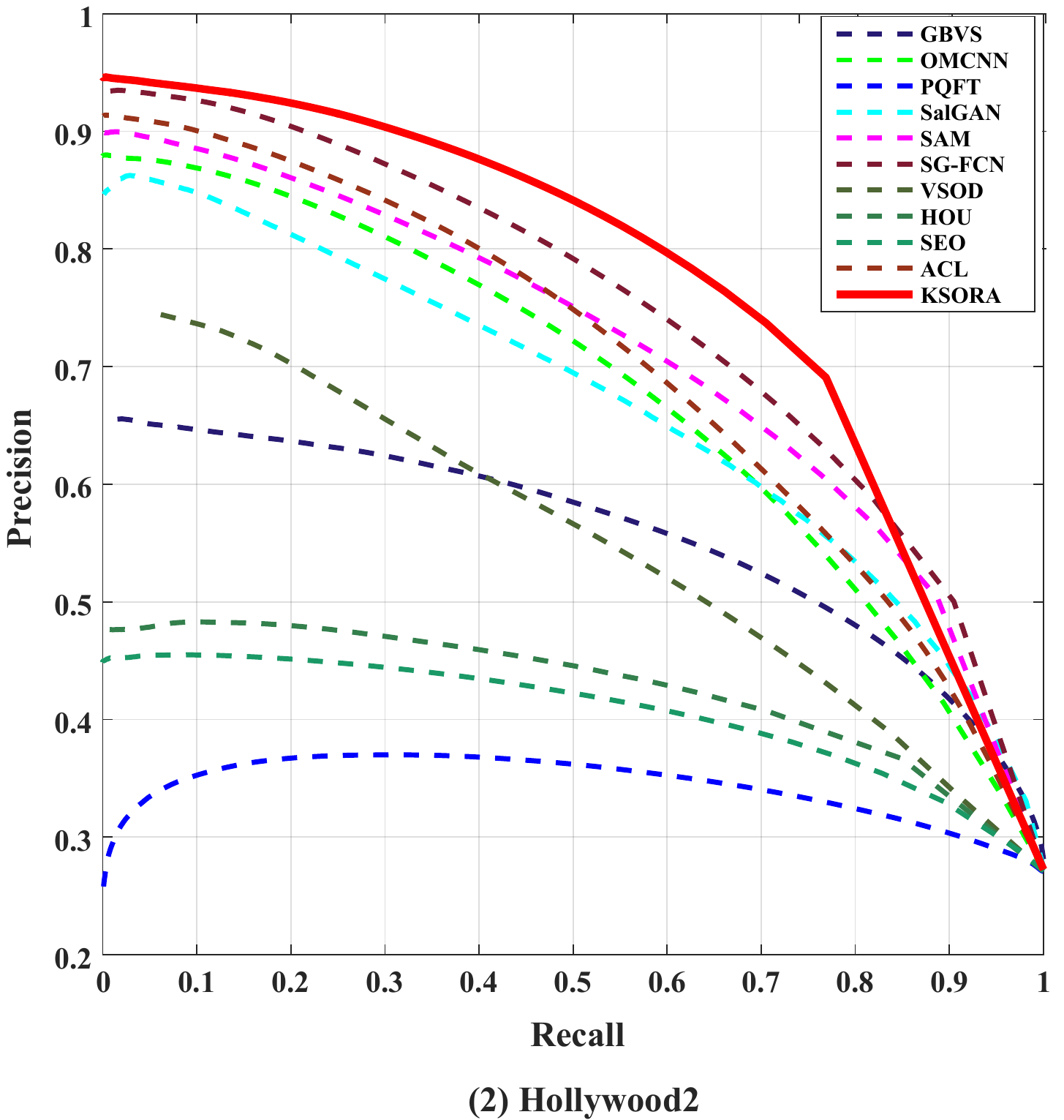}
\end{center}
\caption{Comparison of precision-recall curves of 11 saliency detection methods on UCF and Hollywood2. Our proposed \emph{KSORA} outperforms other competitors across the two testing datasets.}
\label{fig:07}
\end{figure}

To fully verify the performance of the proposed, we evaluate the performance of \emph{KSORA} with ten competitors, which include: SalGAN \cite{Pan_2017_SalGAN}, SAM \cite{8400593}, GBVS \cite{harel2007graph}, PQFT \cite{5223506}, VSOD \cite{8047320}, SG-FCN \cite{8365810}, OMCNN\cite{DBLP:journals/corr/abs-1709-06316}, Hou,$et.al$ \cite{NIPS2008_3531}, Seo, $et.al$ \cite{seo2009static} and ACL \cite{Wang_2018_CVPR_Revisiting}, qualitative and quantitative evaluation results are shown in Table \ref{Tab:01} and Figure \ref{fig:06}-\ref{fig:07}.

From the experimental results shown in the table \ref{Tab:01}, we can see that our model achieve the best results on all three datasets, which proves the effectiveness of our proposed method. In particular, on ucf-sprots, such as the first and the second columns in Figure \ref{fig:06}, when the main object is small and the scene is particularly complex, most models fail to accurately locate the object, some even loss key object, causing saliency detection failure, but our model can accurately fixate the position of the main object and keep the saliency value of the key object at a prominent level. On hollywood2, such as columns 5-6 in Figure \ref{fig:06}, when many objects with equivalent semantic information appear in the scene, most models salient all the objects evenly, while our model can focus on the most significant object, which is more in line with the ground truth samples. On ledov, in addition to the human objects, the main salient objects of this dataset also include animal and other human artifacts, so the scene is more complicated, such as column 3-4 in Figure \ref{fig:06}, our model still ranks first in the overall level and achieve good detection results.
\begin{table*}
\begin{center}
\begin{tabular}{c|c|c|c|c|c|c|c|c|c|c|c|c}
        \thickhline
        \multirow{2}{*}{\footnotesize{Sub-module}}&\multicolumn{4}{c|}{ucf}&\multicolumn{4}{c|}{hollywood}&\multicolumn{4}{c}{ledov}
        \\ \cline{2-13}
        &\footnotesize{CC$\uparrow$}&\footnotesize{SIM$\uparrow$}&\footnotesize{EMD$\downarrow$}&\footnotesize{AUC-J$\uparrow$}&\footnotesize{CC$\uparrow$}&\footnotesize{SIM$\uparrow$}&\footnotesize{EMD$\downarrow$}&\footnotesize{AUC-J$\uparrow$}&\footnotesize{CC$\uparrow$}&\footnotesize{SIM$\uparrow$}&\footnotesize{EMD$\downarrow$}&\footnotesize{AUC-J$\uparrow$}
        \\ \thickhline
        \footnotesize{OBJnet}&0.523&0.326&1.047&0.868&0.462&0.379&1.233&0.839&0.329&0.210&1.163&0.842
        \\
        \footnotesize{OBJ-FLO}&0.515&0.380&0.975&0.873&0.486&0.418&1.152&0.835&0.343&0.240&1.078&0.844
        \\
        \footnotesize{OBJ-FLO-WFE}&0.559&0.389&0.932&0.889&0.510&0.411&1.105&0.852&0.372&0.238&1.088&0.865
        \\
        \footnotesize{KSORA}&\textbf{0.609}&\textbf{0.490}&\textbf{0.747}&\textbf{0.904}&\textbf{0.606}&\textbf{0.508}&\textbf{0.861}&\textbf{0.900}&\textbf{0.496}&\textbf{0.371}&\textbf{0.824}&\textbf{0.899}
        \\ \thickhline    	         																			
\end{tabular}
\end{center}
\caption{Self-ablation experiments. According to the results, it can be concluded that the weighted feature extraction and the key object selection all contribute to the performance improvement of the model.}
\label{Tab:03}
\end{table*}
\subsection{Ablation Study}
Here we perform a self-ablation experiment to further illustrate the role of each module. The config and evaluation results are shown in Table \ref{Tab:03}. According to the results, we can draw the following conclusions: i) The weighted feature extraction module promotes the improvement of model performance by 1-5\% by filtering useless features and emphasizing informative features, ; ii) The key object selection module ranks objects from high semantic level and re-enhances the key salient object region. such operation re-reinforces the features of the key object, and adjust the final saliency prediction of the model to ensure that the saliency value of the key object is maintained at a high level. After deploying \emph{KOS} module, the accuracy of the model has increased by 5-20\%.

\section{Conclusion}\label{Sec.05}
With regard to the dynamic saliency detection problem, we propose a key salient object re-augmentation method \emph{KSORA} based on top-down semantic knowledge and bottom-up feature guidance to improve detection accuracy in video scenes, which is designed to select and enhance the key saliency object in video scenes for more accurate and concentrated saliency detection. Extensive experiments prove the effectiveness of the proposed method.

{\small
\bibliographystyle{ieee}
\bibliography{7437489,Song_2018_CVPR,8100131,Oh_2017_CVPR,Zhu_2018_CVPR,Wang_2018_CVPR,730558,6287326,5963689,7051244,7488288,8382302,8400593,8365810,7164324,Gorji_2018_CVPR,Li_2018_CVPR,8237450,7752954,6618996,8047320,Jiang2017Predicting,8237294,Wang_Lu_2018_CVPR,Wang_2017_CVPR,Zhang_2018_CVPR,Ramanishka_2017_CVPR,Pan_2017_SalGAN,7442536,6751329,Wang_2018_CVPR_Revisiting,YOLO,Flow,8240654,6942210,Soomro2014Action,5223506,8119879,NIPS2008_3531,seo2009static,Hu_2018_CVPR}

\begin{thebibliography}{10}\itemsep=-1pt

\bibitem{8400593}
M.~Cornia, L.~Baraldi, G.~Serra, and R.~Cucchiara.
\newblock Predicting human eye fixations via an lstm-based saliency attentive
  model.
\newblock {\em IEEE Transactions on Image Processing}, 27(10):5142--5154, Oct
  2018.

\bibitem{DBLP:journals/corr/FischerDIHHGSCB15}
P.~Fischer, A.~Dosovitskiy, E.~Ilg, P.~H{\"{a}}usser, C.~Hazirbas, V.~Golkov,
  P.~van~der Smagt, D.~Cremers, and T.~Brox.
\newblock Flownet: Learning optical flow with convolutional networks.
\newblock {\em CoRR}, abs/1504.06852, 2015.

\bibitem{Gorji_2018_CVPR}
S.~Gorji and J.~J. Clark.
\newblock Going from image to video saliency: Augmenting image salience with
  dynamic attentional push.
\newblock In {\em The IEEE Conference on Computer Vision and Pattern
  Recognition (CVPR)}, June 2018.

\bibitem{5223506}
C.~Guo and L.~Zhang.
\newblock A novel multiresolution spatiotemporal saliency detection model and
  its applications in image and video compression.
\newblock {\em IEEE Transactions on Image Processing}, 19(1):185--198, Jan
  2010.

\bibitem{7051244}
J.~Han, D.~Zhang, S.~Wen, L.~Guo, T.~Liu, and X.~Li.
\newblock Two-stage learning to predict human eye fixations via sdaes.
\newblock {\em IEEE Transactions on Cybernetics}, 46(2):487--498, Feb 2016.

\bibitem{harel2007graph}
J.~Harel, C.~Koch, and P.~Perona.
\newblock Graph-based visual saliency.
\newblock In {\em Advances in neural information processing systems}, pages
  545--552, 2007.

\bibitem{5963689}
X.~Hou, J.~Harel, and C.~Koch.
\newblock Image signature: Highlighting sparse salient regions.
\newblock {\em IEEE Transactions on Pattern Analysis and Machine Intelligence},
  34(1):194--201, Jan 2012.

\bibitem{NIPS2008_3531}
X.~Hou and L.~Zhang.
\newblock Dynamic visual attention: searching for coding length increments.
\newblock In D.~Koller, D.~Schuurmans, Y.~Bengio, and L.~Bottou, editors, {\em
  Advances in Neural Information Processing Systems 21}, pages 681--688. Curran
  Associates, Inc., 2009.

\bibitem{Hu_2018_CVPR}
J.~Hu, L.~Shen, and G.~Sun.
\newblock Squeeze-and-excitation networks.
\newblock In {\em The IEEE Conference on Computer Vision and Pattern
  Recognition (CVPR)}, June 2018.

\bibitem{730558}
L.~Itti, C.~Koch, and E.~Niebur.
\newblock A model of saliency-based visual attention for rapid scene analysis.
\newblock {\em IEEE Transactions on Pattern Analysis and Machine Intelligence},
  20(11):1254--1259, Nov 1998.

\bibitem{6751329}
Y.~Jia and M.~Han.
\newblock Category-independent object-level saliency detection.
\newblock In {\em 2013 IEEE International Conference on Computer Vision}, pages
  1761--1768, Dec 2013.

\bibitem{DBLP:journals/corr/abs-1709-06316}
L.~Jiang, M.~Xu, and Z.~Wang.
\newblock Predicting video saliency with object-to-motion {CNN} and two-layer
  convolutional {LSTM}.
\newblock {\em CoRR}, abs/1709.06316, 2017.

\bibitem{Oh_2017_CVPR}
S.~Joon~Oh, R.~Benenson, A.~Khoreva, Z.~Akata, M.~Fritz, and B.~Schiele.
\newblock Exploiting saliency for object segmentation from image level labels.
\newblock In {\em The IEEE Conference on Computer Vision and Pattern
  Recognition (CVPR)}, July 2017.

\bibitem{8237450}
G.~Leifman, D.~Rudoy, T.~Swedish, E.~Bayro-Corrochano, and R.~Raskar.
\newblock Learning gaze transitions from depth to improve video saliency
  estimation.
\newblock In {\em 2017 IEEE International Conference on Computer Vision
  (ICCV)}, pages 1707--1716, Oct 2017.

\bibitem{Li_2018_CVPR}
G.~Li, Y.~Xie, T.~Wei, K.~Wang, and L.~Lin.
\newblock Flow guided recurrent neural encoder for video salient object
  detection.
\newblock In {\em The IEEE Conference on Computer Vision and Pattern
  Recognition (CVPR)}, June 2018.

\bibitem{7488288}
X.~Li, L.~Zhao, L.~Wei, M.~Yang, F.~Wu, Y.~Zhuang, H.~Ling, and J.~Wang.
\newblock Deepsaliency: Multi-task deep neural network model for salient object
  detection.
\newblock {\em IEEE Transactions on Image Processing}, 25(8):3919--3930, Aug
  2016.

\bibitem{6942210}
S.~Mathe and C.~Sminchisescu.
\newblock Actions in the eye: Dynamic gaze datasets and learnt saliency models
  for visual recognition.
\newblock {\em IEEE Transactions on Pattern Analysis and Machine Intelligence},
  37(7):1408--1424, July 2015.

\bibitem{Pan_2017_SalGAN}
J.~Pan, C.~Canton, K.~McGuinness, N.~E. O'Connor, J.~Torres, E.~Sayrol, and
  X.~a. Giro-i Nieto.
\newblock Salgan: Visual saliency prediction with generative adversarial
  networks.
\newblock In {\em arXiv}, January 2017.

\bibitem{Ramanishka_2017_CVPR}
V.~Ramanishka, A.~Das, J.~Zhang, and K.~Saenko.
\newblock Top-down visual saliency guided by captions.
\newblock In {\em The IEEE Conference on Computer Vision and Pattern
  Recognition (CVPR)}, July 2017.

\bibitem{DBLP:journals/corr/RedmonDGF15}
J.~Redmon, S.~K. Divvala, R.~B. Girshick, and A.~Farhadi.
\newblock You only look once: Unified, real-time object detection.
\newblock {\em CoRR}, abs/1506.02640, 2015.

\bibitem{6618996}
D.~Rudoy, D.~B. Goldman, E.~Shechtman, and L.~Zelnik-Manor.
\newblock Learning video saliency from human gaze using candidate selection.
\newblock In {\em 2013 IEEE Conference on Computer Vision and Pattern
  Recognition}, pages 1147--1154, June 2013.

\bibitem{seo2009static}
H.~J. Seo and P.~Milanfar.
\newblock Static and space-time visual saliency detection by self-resemblance.
\newblock {\em Journal of vision}, 9(12):15--15, 2009.

\bibitem{Song_2018_CVPR}
C.~Song, Y.~Huang, W.~Ouyang, and L.~Wang.
\newblock Mask-guided contrastive attention model for person re-identification.
\newblock In {\em The IEEE Conference on Computer Vision and Pattern
  Recognition (CVPR)}, June 2018.

\bibitem{Soomro2014Action}
K.~Soomro and A.~R. Zamir.
\newblock {\em Action Recognition in Realistic Sports Videos}.
\newblock Springer International Publishing, 2014.

\bibitem{8365810}
M.~Sun, Z.~Zhou, Q.~Hu, Z.~Wang, and J.~Jiang.
\newblock Sg-fcn: A motion and memory-based deep learning model for video
  saliency detection.
\newblock {\em IEEE Transactions on Cybernetics}, pages 1--12, 2018.

\bibitem{Wang_2017_CVPR}
L.~Wang, H.~Lu, Y.~Wang, M.~Feng, D.~Wang, B.~Yin, and X.~Ruan.
\newblock Learning to detect salient objects with image-level supervision.
\newblock In {\em The IEEE Conference on Computer Vision and Pattern
  Recognition (CVPR)}, July 2017.

\bibitem{8382302}
L.~Wang, L.~Wang, H.~Lu, P.~Zhang, and X.~Ruan.
\newblock Salient object detection with recurrent fully convolutional networks.
\newblock {\em IEEE Transactions on Pattern Analysis and Machine Intelligence},
  pages 1--1, 2018.

\bibitem{Wang_2018_CVPR}
Q.~Wang, Z.~Teng, J.~Xing, J.~Gao, W.~Hu, and S.~Maybank.
\newblock Learning attentions: Residual attentional siamese network for high
  performance online visual tracking.
\newblock In {\em The IEEE Conference on Computer Vision and Pattern
  Recognition (CVPR)}, June 2018.

\bibitem{Wang_Lu_2018_CVPR}
T.~Wang, L.~Zhang, S.~Wang, H.~Lu, G.~Yang, X.~Ruan, and A.~Borji.
\newblock Detect globally, refine locally: A novel approach to saliency
  detection.
\newblock In {\em The IEEE Conference on Computer Vision and Pattern
  Recognition (CVPR)}, June 2018.

\bibitem{8240654}
W.~Wang and J.~Shen.
\newblock Deep visual attention prediction.
\newblock {\em IEEE Transactions on Image Processing}, 27(5):2368--2378, May
  2018.

\bibitem{Wang_2018_CVPR_Revisiting}
W.~Wang, J.~Shen, F.~Guo, M.-M. Cheng, and A.~Borji.
\newblock Revisiting video saliency: A large-scale benchmark and a new model.
\newblock In {\em The IEEE Conference on Computer Vision and Pattern
  Recognition (CVPR)}, June 2018.

\bibitem{7164324}
W.~Wang, J.~Shen, and L.~Shao.
\newblock Consistent video saliency using local gradient flow optimization and
  global refinement.
\newblock {\em IEEE Transactions on Image Processing}, 24(11):4185--4196, Nov
  2015.

\bibitem{8047320}
W.~Wang, J.~Shen, and L.~Shao.
\newblock Video salient object detection via fully convolutional networks.
\newblock {\em IEEE Transactions on Image Processing}, 27(1):38--49, Jan 2018.

\bibitem{7752954}
T.~Xi, W.~Zhao, H.~Wang, and W.~Lin.
\newblock Salient object detection with spatiotemporal background priors for
  video.
\newblock {\em IEEE Transactions on Image Processing}, 26(7):3425--3436, July
  2017.

\bibitem{7442536}
J.~Yang and M.~Yang.
\newblock Top-down visual saliency via joint crf and dictionary learning.
\newblock {\em IEEE Transactions on Pattern Analysis and Machine Intelligence},
  39(3):576--588, March 2017.

\bibitem{8100131}
Y.~Yu, J.~Choi, Y.~Kim, K.~Yoo, S.~Lee, and G.~Kim.
\newblock Supervising neural attention models for video captioning by human
  gaze data.
\newblock In {\em 2017 IEEE Conference on Computer Vision and Pattern
  Recognition (CVPR)}, pages 6119--6127, July 2017.

\bibitem{Zhang_2018_CVPR}
L.~Zhang, J.~Dai, H.~Lu, Y.~He, and G.~Wang.
\newblock A bi-directional message passing model for salient object detection.
\newblock In {\em The IEEE Conference on Computer Vision and Pattern
  Recognition (CVPR)}, June 2018.

\bibitem{8237294}
P.~Zhang, D.~Wang, H.~Lu, H.~Wang, and B.~Yin.
\newblock Learning uncertain convolutional features for accurate saliency
  detection.
\newblock In {\em 2017 IEEE International Conference on Computer Vision
  (ICCV)}, pages 212--221, Oct 2017.

\bibitem{7437489}
R.~Zhao, W.~Oyang, and X.~Wang.
\newblock Person re-identification by saliency learning.
\newblock {\em IEEE Transactions on Pattern Analysis and Machine Intelligence},
  39(2):356--370, Feb 2017.

\bibitem{Zhu_2018_CVPR}
Z.~Zhu, W.~Wu, W.~Zou, and J.~Yan.
\newblock End-to-end flow correlation tracking with spatial-temporal attention.
\newblock In {\em The IEEE Conference on Computer Vision and Pattern
  Recognition (CVPR)}, June 2018.

\end{thebibliography}
}

\end{document}